\pdfoutput=1

\documentclass[11pt,openany]{article}

\usepackage{naacl2021}

\usepackage{times}
\usepackage{latexsym}
\usepackage{amssymb}
\usepackage{verbatim, listings, framed, tabularx,subfig, fancyvrb}
\usepackage{graphicx, setspace, multirow}
\usepackage{appendix}

\usepackage[T1]{fontenc}
\usepackage[utf8]{inputenc}

\usepackage{microtype}

\title{
Word-Level Alignment of Paper Documents with\\
their Electronic Full-Text Counterparts}

\author{Mark-Christoph M\"uller, Sucheta Ghosh, Ulrike Wittig, and Maja Rey\\
Heidelberg Institute for Theoretical Studies gGmbH, Heidelberg, Germany\\
  \texttt{\{mark-christoph.mueller,sucheta.ghosh,}\\
  \texttt{ulrike.wittig,maja.rey\}@h-its.org}}

\begin{document}
\maketitle
\begin{abstract}
We describe a simple procedure for the automatic creation of word-level alignments between printed documents and their respective full-text versions.
The procedure is unsupervised, uses standard, off-the-shelf components only, and reaches an F-score of $85.01$ in the basic setup and up to $86.63$ when using pre- and post-processing.
Potential areas of application are manual database curation (incl.\ document \emph{triage}) and biomedical expression OCR.
\end{abstract}

\section{Introduction}
\label{sec:intro}
Even though most research literature in the life sciences is \emph{born-digital} nowadays, manual data curation \cite{isfb2018} from these documents still often involves paper. 
For curation steps that require close reading and markup of relevant sections, curators frequently rely on paper printouts and highlighter pens \cite{venkatesan2019}.
Figure \ref{fig:doiscan} shows a page of a typical document used for manual curation.
The potential reasons for this can be as varied as merely sticking to a habit, ergonomic issues related to reading from and interacting with a device, and functional limitations of that device \cite{buchanan2007,koepper2016,clinton2019}. 

\noindent
Whatever the \emph{reason}, the \emph{consequence} is a two-fold media break in many manual curation workflows: first from electronic format (either PDF or full-text XML) to paper, and then back from paper to the electronic format of the curation database.
Given the above arguments in favor of paper-based curation, removing the first media break from the curation workflow does not seem feasible. 
Instead, we propose to bridge the gap between paper and electronic media by automatically creating an alignment between the words on the printed document pages and their counterparts in an electronic full-text version of the same document. 


\noindent
Our approach works as follows:
We automatically create machine-readable versions of printed paper documents (which might or might not contain markup) by scanning them,
applying 
optical character recognition (OCR), and
converting the resulting semi-structured OCR output text 
into a flexible XML format for further processing. 
For this, we use the multilevel XML format of the annotation tool MMAX2\footnote{\url{https://github.com/nlpAThits/MMAX2}} \cite{mueller2006}. 
We retrieve electronic full-text counterparts of the scanned paper documents from PubMedCentral\textsuperscript{\textregistered} in \texttt{.nxml} format\footnote{While PubMedCentral\textsuperscript{\textregistered} is an obvious choice here, other resources with different full-text data formats exist and can also be used. All that needs to be modified is the conversion step (see Section \ref{sec:pmctoxml}).},
and also convert them into MMAX2 format. 
By using a shared XML format for the two heterogeneous text sources, we can capture their content and structural information in a way that provides a \emph{compatible}, though often not identical, word-level tokenization. 
Finally, using a sequence alignment algorithm from bioinformatics and some pre- and post-processing, we create a word-level alignment of both documents. 

\noindent
Aligning words from OCR and full-text documents is challenging for 
several reasons. The OCR output contains various types of \textbf{recognition errors}, many of which involve special symbols, Greek letters like $\mu$ or sub- and superscript characters and numbers, which are particularly frequent in chemical names, formulae, and measurement units, and which are notoriously difficult for OCR \cite{ohyama2019}.

\noindent
If the printed paper document is based on PDF, it usually has an \textbf{explicit page layout}, which is different from the way the corresponding full-text XML document is displayed in a web browser. 
Differences include double- vs.\ single-column layout, but also the way in which tables and figures are rendered and positioned.

\noindent
Finally, printed papers might contain \textbf{additional content} in headers or footers (like e.g.\ download timestamps).
Also, while the references/bibliography section is an integral part of a printed paper and will be covered by OCR, in XML documents it is often structurally kept apart from the actual document text.

\noindent
Given these challenges, attempting data extraction from document \emph{images} if the documents are available in PDF or even full-text format may seem unreasonable.  
We see, however, the following useful applications:

\noindent
\textbf{1. Manual Database Curation}
As mentioned above, manual database curation requires the extraction, normalization, and database insertion of scientific content, often from \emph{paper} documents. 
Given a paper document in which a human expert curator has manually marked a word or sequence of words for insertion into the database, having a \emph{link} from these words to their electronic counterparts can eliminate or at least reduce error-prone and time-consuming steps like manual re-keying.
Also, already existing annotations of the electronic full-text\footnote{Like \url{https://europepmc.org/Annotations}} would also be accessible and could be used to inform the curation decision or to supplement the database entry. 

\noindent
\textbf{2. Automatic PDF Highlighting for Manual \emph{Triage}}
Database curation candidate papers are identified by a process called document \emph{triage} \cite{buchanan2007,hirschman2012} which, despite some attempts towards automation (e.g.\ \newcite{wang2020}), remains a mostly manual process. 
In a nut shell, triage normally involves querying a literature database (like PubMed\footnote{\url{https://pubmed.ncbi.nlm.nih.gov/}}) for specific terms, skimming the list of search results, selecting and skim-reading some papers, and finally downloading and printing the PDF versions of the most promising ones for curation \cite{venkatesan2019}.
Here, the switch from \emph{searching} in the electronic full-text (or abstract) to \emph{printing} the PDF brings about a loss of information, because the terms that 
caused the paper to be retrieved will have to be located again in the print-out. 
A word-level alignment between the full-text and the PDF version would allow to create an enhanced version of the PDF with highlighted search term occurrences \emph{before} printing. 

\noindent
\textbf{3. Biomedical Expression OCR}
Current state-of-the-art OCR systems are very accurate at recognizing standard text using Latin script and baseline typography, but, as already mentioned, they are less reliable for more typographically complex expressions like chemical formulae.
In order to develop specialized OCR systems for these types of expressions, ground-truth data is required in which image regions containing these expressions are labelled with the correct characters and their positional information (see also Section \ref{sec:related}). 
If aligned documents are available, this type of data can easily be created at a large scale.

\vspace{5pt}
\noindent
The remainder of this paper is structured as follows. 
In Section \ref{sec:data}, we 
describe our data set and how it was converted
into the shared XML format.
Section \ref{sec:alignment} deals with the actual alignment procedure, including a description of the optional pre- and post-processing measures. 
In Section \ref{sec:experiments}, we present experiments in which we evaluate the performance of the implemented procedure,
including an ablation of the effects of the individual pre- and post-processing measures.
Quantitative evaluation alone, however, does not convey a realistic idea of the actual usefulness of the procedure, which ultimately needs to be evaluated in the context of real applications including, but not limited to, database curation.
Section \ref{sec:examples}, therefore, briefly presents examples of the alignment and highlighting detection functionality and the biomedical expression OCR use case mentioned above. 
Section \ref{sec:related} discusses relevant related work, and Section \ref{sec:conc} summarizes and concludes the paper with some future work.

\noindent
All the tools and libraries we use are freely available. 
In addition, our implementation can be found at \url{https://github.com/nlpAThits/BioNLP2021}.

\section{Data}
\label{sec:data}
For the alignment of a paper document with its electronic full-text counterpart, what is minimally required is an image of every page of the document, and a full-text XML file of the same document. 
The document images can either be created by scanning or by directly converting the corresponding PDF into an image. 
The latter method will probably yield images of a better quality, because it completely avoids the physical printing and subsequent scanning step, while the output of the former method will be more realistic. We experiment with both types of images (see Section \ref{sec:papertoxml}).
We identify a document by its DOI, and refer to the different versions as DOI$_{xml}$ (from the full-text XML), DOI$_{conv}$, and DOI$_{scan}$. Whenever a distinction between DOI$_{conv}$ and DOI$_{scan}$ is not required, we refer to these versions collectively as DOI$_{ocr}$.

\noindent
Printable PDF documents and their associated \texttt{.nxml} files are readily available at PMC-OAI.\footnote{\url{https://www.ncbi.nlm.nih.gov/pmc/tools/oai/}}
In our case, however, printed paper versions were already available, as we have access to a collection of more than $6.000$ printed scientific papers (approx.\ $30.000$ pages in total) that were created in the SABIO-RK\footnote{\url{http://sabio.h-its.org/}} Biochemical Reaction Kinetics Database project \cite{wittig2017,wittig2018}. 
These papers contain manual highlighter markup at different levels of granularity, including the word, line, and section level. 
Transferring this type of markup from printed paper to the electronic medium is one of the key applications of our alignment procedure. Our paper collection spans many publication years and venues.
For our experiments, however, it was required that each document was freely available both as PubMedCentral\textsuperscript{\textregistered} full-text XML and as PDF. 
While this leaves only a fraction of (currently) $68$ papers, the data situation is still sufficient to demonstrate the feasibility of our procedure. 
Even more importantly, the procedure is unsupervised, i.e.\ it does not involve learning and does not require any training data.

\subsection{Document Image to Multilevel XML}
\label{sec:papertoxml}
\begin{figure*}[t]
\centering
\subfloat[DOI$_{scan}$\label{fig:doiscan}]{
\frame{\includegraphics[width=.3\textwidth]{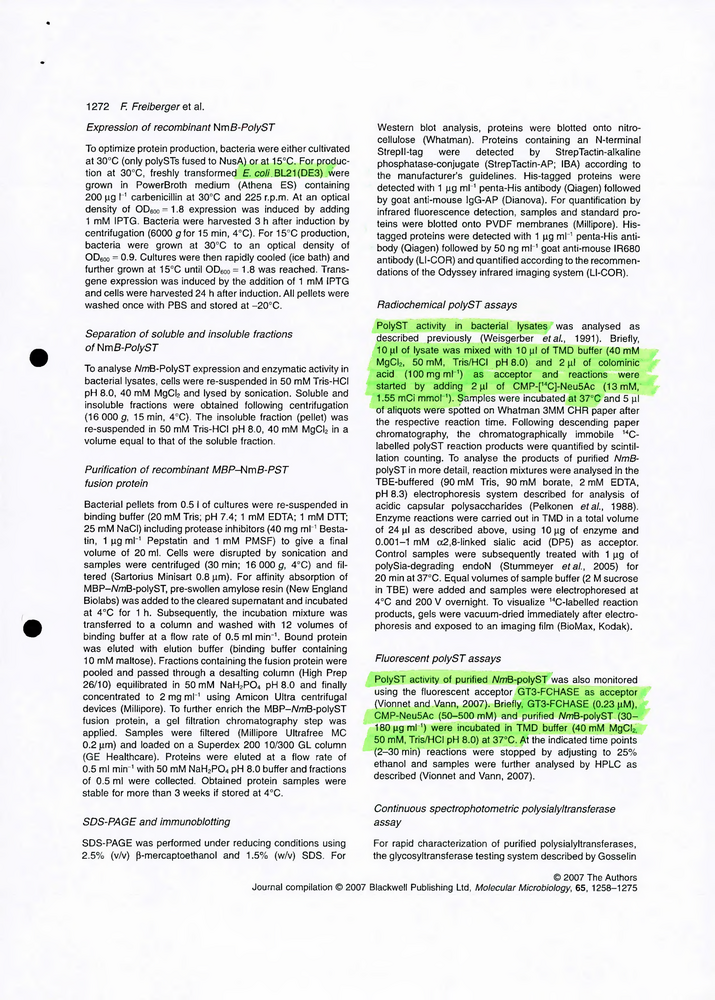}}}
\hfill
\subfloat[DOI$_{scan\_bg}$\label{fig:doiscanbg}]{
\frame{\includegraphics[width=.3\textwidth]{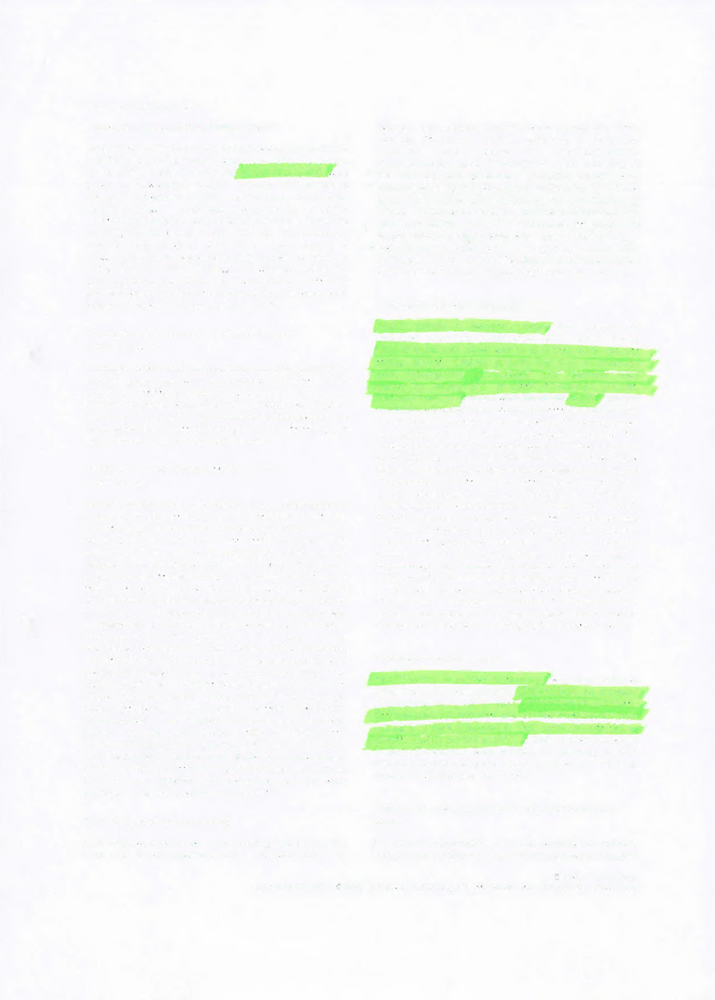}}}
\hfill
\subfloat[DOI$_{conv}$\label{fig:doiconv}]{
\frame{\includegraphics[width=.3\textwidth]{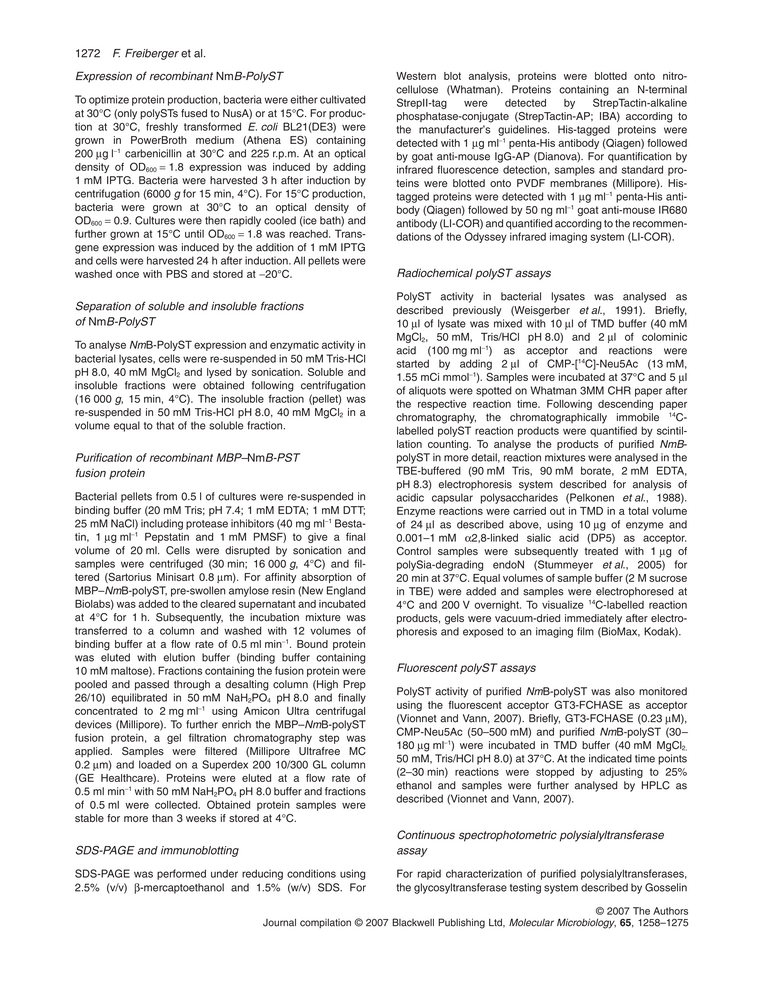}}
}
\caption{Three image renderings of the same document page: Scanned print-out w/ manual markup (a), background with markup only (b), and original PDF (c).}
\label{fig:three_images_per_page}
\end{figure*}
Since we want to compare downstream effects of input images of different quality, we created both a converted and a scanned image version for every document in our data set.
For the DOI$_{conv}$ version, we used \texttt{pdftocairo} to create a high-resolution (600 DPI) PNG file for every PDF page. Figure \ref{fig:doiconv} shows an example.
The DOI$_{scan}$ versions, on the other hand, were extracted from 'sandwich' PDFs which had been created earlier by a professional scanning service provider. 
The choice of a service provider for this task was only motivated by the large number of pages to process, and not by expected quality or other considerations. 
A sandwich PDF contains, among other data, the document plain text (as recognized by the provider's OCR software) and a background image for each page. 
This background image is a by-product of the OCR process in which pixels that were recognized as parts of a character are inpainted, i.e.\ removed by being overwritten with colors of neighbouring regions. 
Figure \ref{fig:doiscanbg} shows the background image corresponding to the page in Figure \ref{fig:doiscan}. Note how the image retains the highlighting.
We used \texttt{pdfimages} to extract the background images (72 DPI) from the sandwich PDF for use in highlighting extraction (see Section \ref{sec:highlighting} below). We refer to these versions as DOI$_{scan\_bg}$.
For the actual DOI$_{scan}$ versions, 
we again used \texttt{pdftocairo} to create a high-resolution (600 DPI) PNG file for every scanned page. 

\noindent
OCR was then performed on the DOI$_{conv}$ and the DOI$_{scan}$ versions with tesseract 4.1.1\footnote{\url{https://github.com/tesseract-ocr/tesseract}}, using default recognition settings (\texttt{--oem 3 --psm 3}) and specifying hOCR\footnote{\url{http://kba.cloud/hocr-spec/1.2/}} with character-level bounding boxes as output format. 
In order to maximize recognition accuracy (at the expense of processing speed), the default language models for English were replaced with optimized LSTM models\footnote{\url{https://github.com/tesseract-ocr/tessdata_best}}.
No other modification or re-training of tesseract was performed. 
In a final step, the hOCR output from both image versions was converted into the MMAX2 \cite{mueller2006} multilevel XML annotation format, using \emph{words} as tokenization granularity, and storing word- and character-level confidence scores and bounding boxes as MMAX2 attributes.\footnote{See the lower part of Figure \ref{fig:datamodel} in the Appendix.}

\subsubsection{Highlighting Detection}
\label{sec:highlighting}
Highlighting detection and subsequent extraction can be performed if the scanned paper documents contain manual markup.
In its current state, the detection procedure described in the following \emph{requires} inpainted OCR background images which, in our case, were produced by the third-party OCR software used by the scanning service provider. 
tesseract, on the other hand, does \emph{not} produce these images. 
While it would be desirable to employ free software only, this fact does not severely limit the usefulness of our procedure, because 1) other software (either free or commercial) with the same functionality might exist, and 2) even for document collections of medium size, employing an external service provider might be the most economical solution even in academic / research settings, anyway.
What is more, inpainted backgrounds are only required if highlighting detection is desired: For text-only alignment, plain scans are sufficient. 

\noindent
The actual highlighting extraction works as follows (see \newcite{mueller2020} for details): 
Since document highlighting comes mostly in strong colors, which are characterized by large differences among their three component values in the RGB color model, we create a binarized version of each page by going over all pixels in the background image and setting each pixel to $1$ if the pairwise differences between the R, G, and B components are above a certain threshold ($50$), and to $0$ otherwise. This yields an image with regions of higher and lower density of black pixels.
In the final step, we iterate over the word-level tokens created from the hOCR output and converted into MMAX2 format earlier, compute for each word its degree of highlighting as the percentage of black pixels in the word's bounding box, and store that percentage value as another MMAX2 attribute if it is at least $50\%$.
An example result will be presented in Section \ref{sec:examples}.

\subsection{PMC\textsuperscript{\textregistered} \texttt{.nxml} to Multilevel XML}
\label{sec:pmctoxml}
The \texttt{.nxml} format employed for PubMedCentral\textsuperscript{\textregistered} full-text documents uses the JATS scheme\footnote{\url{https://jats.nlm.nih.gov/archiving/}} which supports a rich meta data model, only a fraction of which is of interest for the current task. 
In principle, however, all information contained in JATS-conformant documents can also be represented in the multilevel XML format of MMAX2.
The \texttt{.nxml} data provides precise information about both the textual content (including correctly encoded special characters) and its word- and section-level layout.
At present, we only extract content from the \textbf{\texttt{$<$article-meta$>$}} section (\texttt{$<$article-title$>$}, \texttt{$<$surname$>$}, \texttt{$<$given-names$>$}, \texttt{$<$xref$>$}, \texttt{$<$email$>$}, \texttt{$<$aff$>$}, and \texttt{$<$abstract$>$}), and from the 
\textbf{\texttt{$<$body$>$}} (\texttt{$<$sec$>$}, \texttt{$<$p$>$}, \texttt{$<$tr$>$}, \texttt{$<$td$>$}, \texttt{$<$label$>$}, \texttt{$<$caption$>$}, and \texttt{$<$title$>$}).
These sections cover the entire textual content of the document. 
We also extract the formatting tags \texttt{$<$italic$>$}, \texttt{$<$bold$>$}, \texttt{$<$underline$>$}, and in particular \texttt{$<$sup$>$}  and \texttt{$<$sub$>$}. 
The latter two play a crucial role in the chemical formulae and other domain-specific expressions.
Converting the \texttt{.nxml} data to our MMAX2 format is straightforward.\footnote{See the upper part of Figure \ref{fig:datamodel} in the Appendix.}
In some cases, the \texttt{.nxml} files contain embedded LaTex code in \texttt{$<$tex-math$>$} tags. 
If this tag is 
encountered, its content is processed as follows: 
LaTex Math encodings for sub- and superscript, $\string_\{\}$ and $\string^\{\}$, are removed, their content is extracted and re-inserted with JATS-conformant \texttt{$<$sub$>$...$<$/sub$>$} and \texttt{$<$sup$>$...$<$/sup$>$} elements. Then, the resulting LaTex-like string is sent through the \texttt{detex} tool to remove any other markup.
While this obviously cannot handle layouts like e.g.\ fractions, it still preserves many simpler expressions that would otherwise be lost in the conversion. 

\section{Outline of the Alignment Procedure}
\label{sec:alignment}
The actual word-level alignment of the DOI$_{xml}$ version with the DOI$_{ocr}$ version of a document operates on lists of $<token, id>$ tuples which are created from each version's MMAX2 annotation. 
These lists are characterized by longer and shorter stretches of tuples with matching tokens, which just happen to start and end at different list indices. 
These stretches are interrupted at times by (usually shorter) sequences of tuples with non-matching tokens, which mostly exist as the result of OCR errors (see below). 
\emph{Larger} distances between stretches of tuples with matching tokens, on the other hand, can be caused by structural differences between the DOI$_{xml}$ and the DOI$_{ocr}$ version, which can reflect actual layout differences, but which can also result from OCR errors like incorrectly joining two adjacent lines from two columns. 

\noindent
The task of the alignment is to find the correct mapping on the token level for as many tuples as possible. 
We use the \texttt{align.globalxx} method from the \texttt{Bio.pairwise2}
module of Biopython \cite{cock2009}, which provides pairwise sequence alignment using a dynamic programming algorithm \cite{needlemanwunsch1970}.
While this library supports the definition of custom similarity functions for minimizing the alignment cost, we use the most simple version which just applies a binary (=identity) matching scheme, i.e.\ full matches are scored as $1$, all others as $0$.
This way, we keep full control of the alignment, and can identify and locally fix non-matching sequences during post-processing (cf.\ Section \ref{sec:postpro} below). 
The result of the alignment (after optional pre- and post-processing) is an $n$-to-$m$ mapping between $<token, id>$ tuples from the DOI$_{xml}$ and the DOI$_{ocr}$ version of the same document.\footnote{See also the central part of Figure \ref{fig:datamodel} in the Appendix.}

\subsection{Pre-Processing}
\label{sec:prepro}
The main difference between pre- and post-processing is that the former operates on two still \emph{unrelated} tuple lists of different lengths, while for the latter the tuple lists have the same length due to padding entries (\texttt{<<GAP>>}) that were inserted by the alignment algorithm in order to bridge sequences of non-alignable tokens. 
Pre-processing aims to smooth out trivial mismatches and thus to help alignment.
Both pre- and post-processing, however, only modify the tokens in DOI$_{ocr}$, but never those in DOI$_{xml}$, which are considered as gold-standard. 

\noindent
\textbf{Pre-compress matching sequences [\texttt{pre\_compress=\emph{p}}]}
The space complexity of the Needleman-Wunsch algorithm is ${O}(mn)$, where $m$ and $n$ are the numbers of tuples in each document.
Given the length of some documents, the memory consumption of the alignment can quickly become critical.
In order to reduce the number of tuples to be compared, we apply a simple pre-compression step which first identifies sequences of $p$ tuples (we use $p=20$ in all experiments) with \emph{perfectly identical} tokens in both documents, and then replaces them with single tuples where the token and id part consist of concatenations of the individual tokens and ids. 
After the alignment, these compressed tuples are expanded again.

\noindent
While pre-compression was always performed, the pre- and post-processing measures described in the following are optional, and their individual effects on the alignment will be evaluated in Section \ref{sec:quantevaluation}.

\noindent
\textbf{De-hyphenate DOI$_{ocr}$ tokens [\texttt{dehyp}]}
Sometimes, words in the DOI$_{ocr}$ versions are hyphenated due to layout requirements which, in principle, do not exist in the DOI$_{xml}$ versions.
These words appear as three consecutive tuples with either the '-' or '¬' token in the center tuple.  
For de-hyphenation, we search the tokens in the tuple list for DOI$_{ocr}$ for single hyphen characters and reconstruct the potential un-hyphenated word by concatenating the tokens immediately before and after the hyphen. 
If this word exists \emph{anywhere} in the list of DOI$_{xml}$ tokens, we simply substitute the three original $<token,id>$ DOI$_{ocr}$ tuples with one merged tuple. 
De-hyphenation (like all other pre- and post-processing measures) is completely lexicon-free, because the decision whether the un-hyphenated word exists is only based on the content of the DOI$_{xml}$ document.

\noindent
Diverging tokenizations in the DOI$_{xml}$ and DOI$_{ocr}$ document versions are a common cause of local mismatches.
Assuming the tokenization in DOI$_{xml}$ to be correct, tokenizations can be fixed by either joining or splitting tokens in DOI$_{ocr}$.

\noindent
\textbf{Join incorrectly split DOI$_{ocr}$ tokens [\texttt{pre\_join}]}
We apply a simple rule to detect and join tokens that were incorrectly split in DOI$_{ocr}$.
We move a window of size $2$ over the list of DOI$_{ocr}$ tuples and concatenate the two tokens. 
We then iterate over all tokens in the DOI$_{xml}$ version.
If we find the reconstructed word in a matching context
(one immediately preceeding and following token), we replace, in the DOI$_{ocr}$ version, the first original tuple with the concatenated one, assigning the concatenated ID as new ID, and remove the second tuple from the list.  Consider the following example. 
\begin{lstlisting}[basicstyle=\small]]
$<$ phen,    word_3084 $>_n$
$<$ yl,      word_3085 $>_{n+1}$
$\Longrightarrow$
$<$ phenyl, word_3084+word_3085 $>_n$
\end{lstlisting}
This process (and the following one) is repeated until no more modifications can be performed.

\noindent
\textbf{Split incorrectly joined DOI$_{ocr}$ tokens [\texttt{pre\_split}]} 
In a similar fashion, we identify and split incorrectly joined tokens.
We move a window of size $2$ over the list of DOI$_{xml}$ tuples, concatenate the two tokens, and try to locate a corresponding single token, in a matching context, in the list of DOI$_{ocr}$ tuples. 
If found, we replace the respective tuple in that list with two new tuples, one with the first token from the DOI$_{xml}$ tuple and one with the second one. 
Both tuples retain the ID from the original DOI$_{ocr}$ tuple. In the following example, the correct tokenization separates the trailing number $3$ from the rest of the expression, because it needs to be typeset in subscript in order for the formula to be rendered correctly. 
\begin{lstlisting}[basicstyle=\small]]
$<$ KHSO3,   word_3228 $>_n$
$\Longrightarrow$
$<$ KHSO,    word_3228 $>_n$
$<$ 3,       word_3228 $>_{n+1}$
\end{lstlisting}

\subsection{Alignment Post-Processing}
\label{sec:postpro}
\textbf{Force-align [\texttt{post\_force\_align}]}
The most frequent post-processing involves cases where single tokens of the same length and occurring in the same context are not aligned automatically.
In the following, the left column contains the DOI$_{ocr}$ and the right the DOI$_{xml}$ tuples. In the first example, the $\beta$ was not correctly recognized and substituted with a B. 
We identify force-align candidates like these by looking for sequences of $s$ consecutive tuples with a <<GAP>> token in one list, followed by a similar sequence of the same length in the other. Then, if both the context and the number of characters matches, we force-align the two sequences.
\begin{lstlisting}[basicstyle=\scriptsize]]
$<$metallo, word_853$>$      $<$metallo, word_546$>$ 
$<$-, word_854$>$            $<$-, word_547$>$
$<$B, word_855$>$            $<$<<GAP>>, -$>$
$<$<<GAP>>, -$>$             $<\beta$, word_548$>$
$<$-, word_856$>$            $<$-, word_549$>$
$<$lactamase, word_857$>$  $<$lactamase, word_550$>$
$\Longrightarrow$
...
$<$B, word_855$>$            $<\beta$, word_548$>$
...
\end{lstlisting}

\noindent
For $s=2$, force-align will also fix the following.
\begin{lstlisting}[basicstyle=\scriptsize]]
$<$acid, word_1643$>$        $<$acid, word_997$>$ 
$<$,, word_1644$>$           $<$,, word_998$>$
$<$1t, word_1645$>$          $<$<<GAP>>, -$>$
$<$1s, word_1646$>$          $<$<<GAP>>, -$>$
$<$<<GAP>>, -$>$             $<$it, word_999$>$
$<$<<GAP>>, -$>$             $<$is, word_1000$>$
$<$purified, word_1647$>$    $<$purified, word_1001$>$
$<$using, word_1648$>$       $<$using, word_1002$>$
$\Longrightarrow$
...
$<$1t, word_1645$>$          $<$it, word_999$>$
$<$1s, word_1646$>$          $<$is, word_1000$>$
...
\end{lstlisting}

\section{Experiments}
\label{sec:experiments}
\subsection{Quantitative Evaluation}
\label{sec:quantevaluation}
\begin{table*}[htbp]
\centering
\begin{small}
\begin{tabular}{l|l|l|l|l|l|l|}
\multicolumn{1}{c|}{\multirow{2}{*}{\textbf{Pre-/Post-Processing}}}  & \multicolumn{3}{c|}{\textbf{DOI$_{xml}$ -- DOI$_{conv}$}} & \multicolumn{3}{c|}{\textbf{DOI$_{xml}$ -- DOI$_{scan}$}} \\ 
    &                        \multicolumn{1}{c}{P} & \multicolumn{1}{c}{R} & \multicolumn{1}{c|}{F} & \multicolumn{1}{c}{P} & \multicolumn{1}{c}{R} & \multicolumn{1}{c|}{F} \\ 
\hline
\texttt{-}                                  & $\textbf{95.04}$ & $76.90$ & $85.01$ & $\textbf{93.59}$ & $75.29$ & $83.45$ \\
\texttt{dehyp}                              & $94.91$ & $77.47$ & $85.31$ & $93.48$ & $75.96$ & $83.81$ \\
\texttt{pre}                                & $\textbf{95.04}$ & $77.40$ & $85.32$ & $93.57$ & $75.83$ & $83.77$ \\
\texttt{dehyp + pre}                        & $94.90$ & $77.97$ & $85.61$ & $93.47$ & $76.52$ & $84.15$ \\
\texttt{post\_force\_align}                 & $95.03$ & $78.57$ & $86.02$ & $93.57$ & $76.99$ & $84.48$ \\
\texttt{dehyp + post\_force\_align}         & $94.91$ & $79.17$ & $86.32$ & $93.47$ & $77.69$ & $84.86$ \\
\texttt{pre + post\_force\_align}           & $95.02$ & $79.08$ & $86.32$ & $93.56$ & $77.55$ & $84.81$ \\
\texttt{dehyp + pre + post\_force\_align}   & $94.90$ & $\textbf{79.68}$ & $\textbf{86.63}$ & $93.47$ & $\textbf{78.27}$ & $\textbf{85.20}$ \\
\hline                                      \end{tabular}
\end{small}
\caption{Alignment Scores (micro-averaged, n=$68$). All results using \texttt{pre\_compress=$20$}.
Max.\ values in bold.}
\label{tab:alignment_scores_overview}
\end{table*}
We evaluate the system on our $68$ DOI$_{xml}$ -- DOI$_{ocr}$
document pair data set by computing P, R, and F for the task of aligning tokens from DOI$_{xml}$ (the gold-standard) to tokens in DOI$_{ocr}$.
By defining the evaluation task in this manner, we take into account that the DOI$_{ocr}$ version usually contains more tokens, mostly because it includes the bibliography, which is generally \emph{not} included in the DOI$_{xml}$ version.
Thus, an alignment is perfect if every token in DOI$_{xml}$ is correctly aligned to a token in DOI$_{ocr}$, regardless of there being \emph{additional} tokens in DOI$_{ocr}$.
\noindent
In order to compute P and R, the number of \emph{correct} alignments (=TP) among all alignments needs to be determined.
Rather than inspecting and checking all alignments manually, we employ a simple heuristic: 
Given a pair of automatically aligned tokens,
we create two KWIC string representations, KWIC$_{xml}$ and KWIC$_{ocr}$, with a left and right context of $10$ tokens each.
Then, we compute the normalized Levenshtein similarity $lsim$ between each pair $ct1$ and $ct2$ of left and right contexts, respectively, as 
\[{1-levdist(ct1, ct2)/max(len(ct1), len(ct2))}\]
\noindent
We count the alignment as correct (=TP) if $lsim$ of \textbf{both} the two left \textbf{and} the two right contexts is $>=.50$, and as incorrect (=FP) otherwise.\footnote{Figure \ref{fig:kwic} in the Appendix provides an example.}
The number of missed alignments (=FN) can be computed by substracting the number of TP from the number of all tokens in DOI$_{xml}$. 
Based on these counts, we compute precision (P), recall (R), and F-score (F) in the standard way.
Results are provided in Table \ref{tab:alignment_scores_overview}. 
For each parameter setting (first column), there are two result columns with P, R, and F each.
The column \textbf{DOI$_{xml}$ -- DOI$_{conv}$} contains alignment results for which OCR was performed on the converted PDF pages, while results in column \textbf{DOI$_{xml}$ -- DOI$_{scan}$} are based on scanned print-outs.
Differences between these two sets of results are due to the inferior quality of the images used in the latter.
The top row in Table \ref{tab:alignment_scores_overview} contains the result of using only the alignment without any pre- or post-processing.
Subsequent rows show results for all possible combinations of pre- and post-processing measures (cf.\ Section \ref{sec:prepro}). 
Note that \texttt{pre\_split} and \texttt{pre\_join} are not evaluated separately and appear combined as \texttt{pre}. 
The first observation is that, for \textbf{DOI$_{xml}$ -- DOI$_{conv}$} and \textbf{DOI$_{xml}$ -- DOI$_{scan}$}, precision is very high, with max.\ values of $95.04$ and $93.59$, respectively.
This is a result of the rather strict alignment method which will align two tokens only if they are \emph{identical} (rather than merely \emph{similar}).
At the same time, precision is very stable across experiments, i.e.\ indifferent to changes in pre- and post-processing. 
This is because, as described in Section \ref{sec:prepro}, pre- and post-processing exclusively aim to improve recall by either smoothing out trivial mismatches before alignment, or adding missing alignments afterwards. 
In fact, pre- and post-processing actually \emph{introduce} precision errors, since they relax this alignment condition somewhat:
This is evident in the fact that the two top precision scores result from the setup with no pre- or post-processing at all, and even though the differences across experiments are extremly small, the pattern is still clear. 
Table \ref{tab:alignment_scores_overview} also shows the intended positive effect of the different pre- and post-processing measures on recall. 
Without going into much detail, we can state the following:
For \textbf{DOI$_{xml}$ -- DOI$_{conv}$} and \textbf{DOI$_{xml}$ -- DOI$_{scan}$}, the lowest recall results from the setup without pre- or post-processing. 
When pre- and post-processing measures are added, recall increases constantly, at the expense of small drops in precision.
However, the positive effect consistently outweighs the negative, causing the F-score to increase to a max.\ score of $86.63$ and $85.20$, respectively, when all pre- and post-processing measures are used. 
Finally, as expected, the inferior quality of the data in DOI$_{scan}$ as compared to DOI$_{conv}$ is nicely reflected in consistently lower scores across all measurements. 
The absolute differences, however, are very small, amounting to only about $1.5$ points. 
This might be taken to indicate that converted (rather than printed and scanned) PDF documents can be functionally equivalent as input for tasks like OCR ground-truth data generation.

\subsection{Qualitative Evaluation and Examples}
\label{sec:examples}
\begin{figure*}[h]
\centering
\frame{\begin{tabular}{lr}
\includegraphics[scale=0.45]{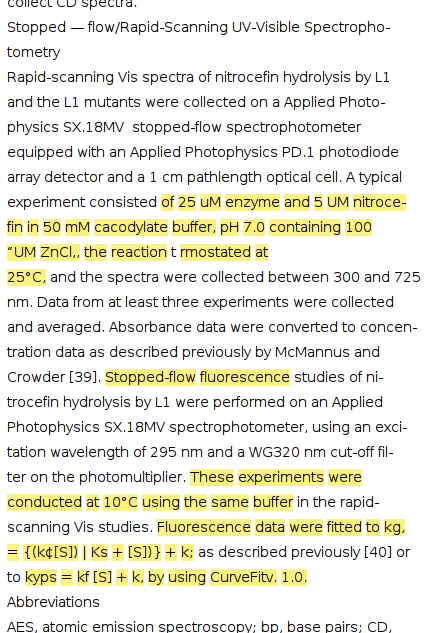}
&
\includegraphics[scale=0.45]{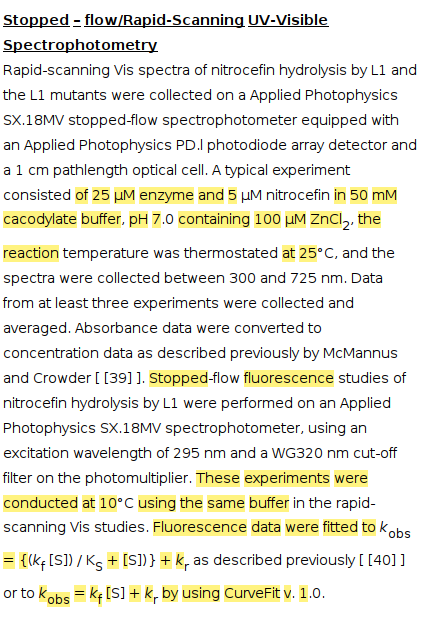}
\end{tabular}}
\caption{DOI$_{scan}$ document with automatically detected, overlayed highlighting (left). DOI$_{xml}$ with highlighting transferred from automatically aligned DOI$_{scan}$ document (right).}
\label{fig:mmax_shots}
\end{figure*}

\noindent
This section complements the quantitative evaluation with some illustrative examples. 
Figure \ref{fig:mmax_shots} shows two screenshots in which 
DOI$_{scan}$ (left) and DOI$_{xml}$ (right) are displayed in the MMAX2 annotation tool. 
\noindent
The left image shows that the off-the-shelf text recognition accuracy of tesseract is very good for standard text, but lacking, as expected, when it comes to recognising special characters and subscripts (like $\mu$, ZnCl$_2$, or k$_{obs}$ in the example).
For the highlighting detection, the yellow text background was chosen as visualization in MMAX2 in order to mimick the physical highlighting of the printed paper. 
Note that since the highlighting detection is based on layout position only (and not anchored to text), manually highlighted text is recognized as highlighted regardless of whether the actual underlying text is recognized correctly. 
\noindent
The right image shows the rendering of the correct text extracted from the original PMC\textsuperscript{\textregistered} full-text XML. 
The rendering of the title as bold and underlined is based on typographic information that was extracted at conversion time (cf.\ Section \ref{sec:pmctoxml}).
The same is true for the subscripts, which are correctly rendered both in terms of the content and the position.
Table \ref{tab:ocr_shots} displays a different type of result, i.e.\ a small selection of a much larger set of OCR errors with their respective images and the correct recognition result. This data, automatically identified by the alignment post-processing, is a valuable resource for the development of biomedical expression OCR systems.

\begin{table}[htbp]
\begin{tabular}{c|c|c}
\setlength\tabcolsep{8.0pt} 
\textbf{Image} & \textbf{OCR} & \textbf{PMC\textsuperscript{\textregistered}}\\
\hline
$\vcenter{\hbox{\includegraphics[scale=0.30]{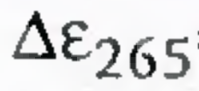}}}$ & \huge{Agyg5} & \huge{$\Delta\varepsilon_{\large{265}}$}\\
\hline
$\vcenter{\hbox{\includegraphics[scale=0.28]{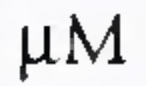}}}$ & \huge{“UM} & \huge{$\mu$M}\\
\hline
$\vcenter{\hbox{\includegraphics[scale=0.28]{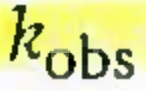}}}$ & \huge{kyps} & \huge{k$_{obs}$}\\
\hline
$\vcenter{\hbox{\includegraphics[scale=0.28]{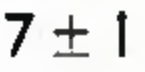}}}$ & \huge{7 + 1} & \huge{$7 \pm 1$}\\
\hline
$\vcenter{\hbox{\includegraphics[scale=0.28]{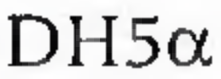}}}$ & \huge{DH50} & \huge{DH5$\alpha$}\\
\hline
\end{tabular}
\caption{Some examples of image snippets (left) with incorrect (middle) and correct (right) text representation.}
\label{tab:ocr_shots}
\end{table}
\section{Related Work}
\label{sec:related}
The work in this paper is obviously related to automatic text alignment, with the difference that what is mostly done there is the alignment of texts in different languages (i.e.\ \textbf{bi-lingual alignment}). 
\newcite{gale1993} align not words but entire sentences from two languages based on statistical properties.
Even if words were aligned, alignment candidates in bi-lingual corpora are not identified on the basis of simple matching, with the exception of language-independent tokens like e.g.\ proper names.

\noindent
Scanning and OCR is also often applied to historical documents, which are only available in paper \cite{hill2019,strien2020,schaefer2020}.
Here, \textbf{OCR post-correction} attempts to map words with word- and character-level OCR errors (similar to those found in our DOI$_{ocr}$ data) to their correct variants, but it does so by using general language models and dictionaries, and not an aligned correct version. 
Many of the above approaches have in common that they employ specialized OCR models and often ML/DL models of considerable complexity. 

\noindent
The idea of using an electronic and a paper version of \textbf{the same document} for creating a character-level alignment dates back at least to \newcite{kanungo1999}, who worked on OCR ground-truth data generation. 
Like most later methods, the procedure of \newcite{kanungo1999} works on the \emph{graphical} level, as opposed to the \emph{textual} level. 
\newcite{kanungo1999} use LaTex to create what they call 'ideal document images' with controlled content.
Print-outs of these images are created, which are then photocopied and scanned, yielding slightly noisy and skewed variants of the 'ideal' images. 
Then, corresponding feature points in both images are identified, and a projective transformation between these is computed.
Finally, the actual ground-truth data is generated by applying this transformation for aligning the bounding boxes in the ideal images 
to their correspondences in the scanned images. 
Since \newcite{kanungo1999} have full control over the content of their 'ideal document images', extracting the ground-truth character data is trivial.
The approach of \newcite{beusekom2008} is similar to that of \newcite{kanungo1999}, but the former use more sophisticated methods, including Canny edge detection \cite{canny1986} for finding corresponding sections in images of the original and the scanned document, and RAST \cite{breuel2001} for doing the actual alignment. 
Another difference is that \newcite{beusekom2008} use pre-existing PDF documents as the source documents from which ground-truth data is to be extracted. 
Interestingly, however, 
their experiments only use synthetic ground-truth data from the UW3 data set\footnote{\url{http://tc11.cvc.uab.es/datasets/DFKI-TGT-2010_1}}, in which bounding boxes and the contained characters are explicitly encoded.
In their conclusion, \newcite{beusekom2008} concede that extracting ground-truth data from PDF is a non-trivial task in itself.
\newcite{ahmed2016} work on automatic ground-truth data generation for \emph{camera-captured} document images, which they claim pose different problems than document images created by scanning, like e.g.\ blur, perspective distortion, and varying lighting. 
Their procedure, however, is similar to that of \newcite{beusekom2008}.
They also use pre-existing PDF documents and automatically rendered 300 DPI images of these documents. 

\noindent

\section{Conclusions}
\label{sec:conc}
In this paper, we described a completely unsupervised procedure for automatically aligning printed paper documents with their electronic full-text counterparts. 
Our point of departure and main motivation was the idea to alleviate the effect of the \textbf{paper-to-electronic media break} in manual biocuration, where printed paper is still very popular when it comes to close reading and manual markup. 
We also argued that the related task of document \emph{triage} can benefit from the availability of alignments between electronic full-text documents (as retrieved from a literature database) and the corresponding PDF documents. 
Apart from this, we identified yet another field of application, biomedical expression OCR, which can benefit from ground-truth data which can automatically be generated with our procedure. 
Improvements in biomedical expression OCR, then, can feed back into the other use cases, by improving the OCR step and thus the alignment, thus potentially establishing a kind of bootstrapping development.
Our implementation relies on \emph{tried and tested} technology, including tesseract as off-the-shelf OCR component, Biopython for the alignment, and MMAX2 as visualization and data processing platform.
The most computationally complex part is the actual sequence alignment with a dynamic programming algorithm from the Biopython library, which we keep tractable even for longer documents by using a simple pre-compression method. 
\noindent
The main experimental finding of this paper is that our approach, although very simple, yields a level of performance that we consider suitable for practical applications.
In quantitative terms, the procedure reaches a very good F-score of $86.63$ on converted and $85.20$ on printed and scanned PDF documents, with corresponding precision scores of $94.90$ and $93.47$, respectively. The negligible difference in results between the two types of images is interesting, as it seems to indicate that converted PDF documents, which are very easy to generate in large amounts, are almost equivalent to the more labour-intensive scans.
In future work, we plan to implement solutions for the identified use cases, and to test them in actual biocuration settings.
Also, we will start creating OCR ground-truth data at a larger scale, and apply that for the development of specialised tools for biomedical OCR. 
\noindent
In the long run, procedures like the one presented in this paper might contribute to the development of systems that support curators to work in a more natural, practical, convenient, and efficient way.

\section*{Acknowledgements}
This work was done as part of the project DeepCurate, which is funded by the German Federal Ministry of Education and Research (BMBF) (No.\ 031L0204) and the Klaus Tschira Foundation, Heidelberg, Germany. We thank the anonymous BioNLP reviewers for their helpful suggestions.




\bibliographystyle{acl_natbib}
\bibliography{custom}

\begin{appendices}
\section*{Appendix}
\renewcommand{\thefigure}{A.\arabic{figure}}
\setcounter{figure}{0}
\begin{figure*}[htbp]
\centering
\includegraphics[width=1.0\textwidth]{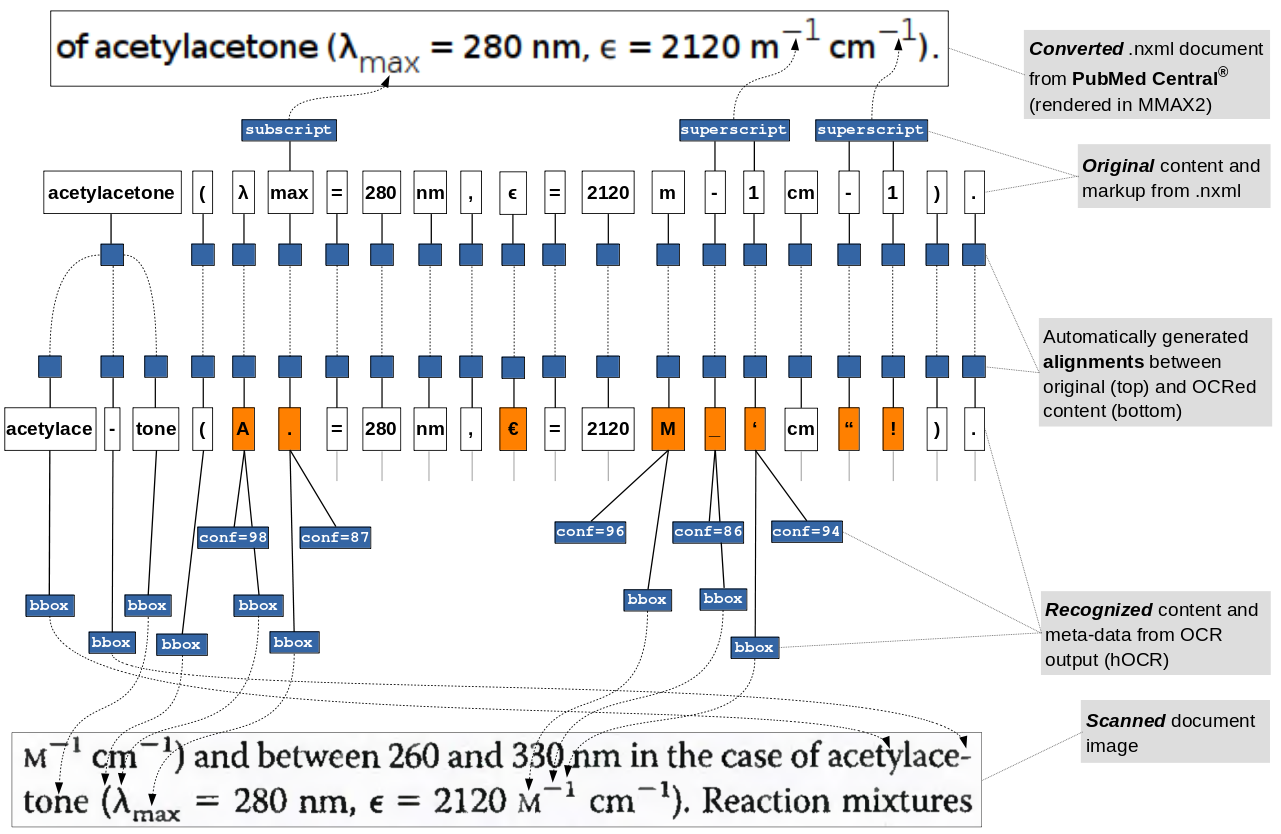}
\caption{Conversion and alignment data model. \textbf{Top}: Full-text and markup (\texttt{subscript}, \texttt{superscript}) is extracted from .nxml documents. Each content token is associated with an alignment token (solid blue boxes). \textbf{Bottom}: Text and meta-data is extracted from the OCR result of scanned document pages. Meta-data includes bounding boxes, which link the recognized text to image regions, and numerical recognition scores, which reflect the confidence with which the OCR system recognized the respective token. (Not all meta-data is given in the Figure to avoid clutter.)}
\label{fig:datamodel}
\end{figure*}

\begin{figure*}
\begin{lstlisting}[basicstyle=\fontsize{5}{6}\selectfont\ttfamily]]
                              MUT) DeficiencyFornyPatrick  1  2  $\dagger$ FroeseD  >>  $\textbf{.}$  <<    Sean  1  3  $\dagger$ SuormalaTerttu  1 YueWyatt W
                                    .org 1,2% Communicated by Elizabeth F  >>  $\textbf{.}$  <<    L Neufeld Received 6 June 2014; accepted revised manuscript
FP (0.136, 0.186)

                                Sean  1  3  $\dagger$ SuormalaTerttu  1 YueWyatt W  >>  $\textbf{.}$  <<     3 BaumgartnerMatthias R.  1  21 Division for
                    6 June 2014; accepted revised manuscript 27 July 2014  >>  $\textbf{.}$  <<    ABSTRACT: Methylmalonyl-CoA mutase (MUT) is
FP (0.208, 0.068)

  ZurichSwitzerland3 Structural Genomics ConsortiumUniversity of OxfordUK  >>  $\textbf{ABSTRACT}$  <<   Methylmalonyl$‐$CoA mutase (MUT) is
                     June 2014; accepted revised manuscript 27 July 2014.  >>  $\textbf{ABSTRACT}$  <<   : Methylmalonyl-CoA mutase (MUT) is an
FP (0.169, 0.842)

Switzerland3 Structural Genomics ConsortiumUniversity of OxfordUKABSTRACT  >>  $\textbf{Methylmalonyl}$  <<   $‐$CoA mutase (MUT) is an
                    ; accepted revised manuscript 27 July 2014. ABSTRACT:  >>  $\textbf{Methylmalonyl}$  <<   -CoA mutase (MUT) is an essential enzyme
FP (0.247, 0.550)

                          University of OxfordUKABSTRACTMethylmalonyl$‐$CoA  >>   $\textbf{mutase}$  <<    (MUT) is an essential enzyme in propionate catabolism
                     manuscript 27 July 2014. ABSTRACT: Methylmalonyl-CoA  >>  $\textbf{mutase}$  <<    (MUT) is an essential enzyme in propionate catabolism
TP (0.538, 1.000)

                              of OxfordUKABSTRACTMethylmalonyl$‐$CoA mutase  >>   $\textbf{(}$  <<   MUT) is an essential enzyme in propionate catabolism that
                         27 July 2014. ABSTRACT: Methylmalonyl-CoA mutase  >>  $\textbf{(}$  <<   MUT) is an essential enzyme in propionate catabolism that
TP (0.667, 1.000)

                               OxfordUKABSTRACTMethylmalonyl$‐$CoA mutase (  >>   $\textbf{MUT}$  <<   ) is an essential enzyme in propionate catabolism that requires
                          July 2014. ABSTRACT: Methylmalonyl-CoA mutase (  >>  $\textbf{MUT}$  <<   ) is an essential enzyme in propionate catabolism that requires
TP (0.702, 1.000)
\end{lstlisting}
\caption{Sample output of the KWIC-based alignment evaluation procedure (context size = $10$ tokens (left and right)). 
In each pair of lines, the top line comes from DOI$_{ocr}$, the bottom line from DOI$_{xml}$. 
Pairs are labeled as TP (correctly aligned) if the normalized Levenshtein similarity of both the left and the right context strings (given in parentheses) is above $0.5$.}
\label{fig:kwic}
\end{figure*}
\end{appendices}

\end{document}